\documentclass[conference]{IEEEtran}
\IEEEoverridecommandlockouts

\usepackage{cite}
\usepackage{amsmath,amssymb,amsfonts}
\usepackage{algorithmic}
\usepackage{graphicx}
\usepackage{textcomp}
\usepackage{xcolor}

\usepackage{pgfplots}
\pgfplotsset{compat=1.18}

\usepackage{booktabs}
\usepackage{multirow}
\usepackage{subcaption}
\usepackage{array}
\usepackage{url}
\usepackage{tabularx}
\usepackage{tikz}
\usetikzlibrary{positioning,shapes.geometric,arrows.meta}

\def\BibTeX{{\rm B\kern-.05em{\sc i\kern-.025em b}\kern-.08em
    T\kern-.1667em\lower.7ex\hbox{E}\kern-.125emX}}

\begin{document}

\title{A 3D Pose-Based Ensemble Framework for Cricket Shot Classification and Automated Biomechanical Analysis}

\author{
\IEEEauthorblockN{
Sourav Shome,
M.d. Ashiquzzaman Rahad,
Dr. Rameswar Debnath
}   
\IEEEauthorblockA{
\textit{Computer Science and Engineering Discipline}\\
\textit{Khulna University}\\
sourav2117@cseku.ac.bd,
rahad.cseku@gmail.com,
rdebnath@cse.ku.ac.bd
}
}
\maketitle

\begin{abstract}
Cricket is one of the most celebrated sports worldwide, and technological advancement has become deeply embedded in how the modern game is analyzed and coached. Cricket shot classification and automated performance analysis add a further dimension to this trend. Traditional approaches rely on RGB video features or static images, which are sensitive to environmental variations such as camera angle, lighting, and background clutter, and often fail to capture the underlying biomechanics of batting actions. In this paper, we propose a system to improve cricket coaching that takes raw video data, extracts batsman from video frames using YOLO, and extracts 3D pose data from video frames using MeTRAbs. The system produces sequential skeletal pose data of 30 body points and captures the biomechanical features of a batsman. As part of the system, we also propose a deep learning ensemble for shot classification of four shots: flick, pull, defense, and drive. The ensemble performed well, compared to existing classification works, achieving 97.68\% accuracy. In addition, we analyzed the misclassification rates to identify cases where shots were incorrectly classified and examined their possible causes. Our proposed system allows novice players to obtain useful feedback, such as important joint angles relative to expert batsmen, which can also be useful for injury prevention. The shot classifier also helps track class-wise shots over time for further analysis. In addition to novice players, coaches can use the system for player evaluation.


\end{abstract}


\section{Introduction}
Cricket is among the most-watched sports worldwide, and success in it is highly dependent on disciplined batting technique. Recent work has begun to apply AI, machine learning, and computer vision to automate cricket performance analysis and support player development. In order to improve the batting techniques, a batsman needs to assess his/her shots. One of the aspects of this assessment is the classification of shots played by players. Most existing shot-classification approaches rely on RGB video features, which are sensitive to camera angle, occlusion, and lighting, and often miss subtle biomechanical variations that distinguish visually similar shots.

Several studies have been done on cricket shot classification using different types of input data, including images, raw videos, and human pose information. Image-based approaches using CNNs and Vision Transformers have achieved accuracies of around 91--93\% \cite{A.S2023CRICKET}, \cite{Dey2024Shot-ViT}, \cite{Fernandes2023Cricket} . Video-based methods, including LRCN and other deep learning approaches, have also been investigated, with reported accuracies ranging from 73\% to 93\% \cite{kanagal2025cricnet}, \cite{10441573}. More recent studies have incorporated 2D pose estimation to capture body movements and improve shot classification \cite{Dey2024Shot-ViT}, \cite{sen2021cricshotclassify}. Comparatively, limited work has been done on 3D pose-based approaches for cricket shot classification, despite their potential to capture more detailed spatial and temporal motion patterns \cite{siddiqui2023enhancing}. This highlights an opportunity for further research on 3D pose-based cricket shot analysis.

In this work, we propose a 3D pose-based framework for cricket shot classification and coaching. The framework processes raw video, detects the batsman using YOLO, and extracts sequential 3D pose data from 30 body points using MeTRAbs to represent the spatio-temporal and biomechanical aspects of batting movements. We evaluate both traditional machine learning and deep learning models designed for sequential data and develop a deep learning ensemble for classifying four shots: flick, pull, defense, and drive, achieving 97.68\% accuracy. The framework also provides joint-angle comparisons with expert batsmen, supporting player feedback, shot tracking, and further performance analysis.

The major contributions of this paper are as follows:

\begin{enumerate}

\item We built a pipeline to convert raw 2D video data into 3D joint coordinates while keeping only the striker batsman and eliminating other noise.

\item We developed a 3d pose-based classification model, evaluating four main types of cricket shots (drives, pulls, flicks, and defensive strokes).

\item We also built a tool that compares the skeletons of two different shot attempts and highlights which joints moved differently.

\end{enumerate}
The rest of the paper is organized as follows. Section II reviews existing approaches to cricket shot classification. Section III presents the proposed methodology, including dataset, preprocessing and classification techniques. Section IV presents and analyzes the results, including baseline comparisons and error analysis. Section V concludes the paper and discusses future work.

\section{Literature Review}

Researchers have been working on Cricket shot identification extensively utilizing deep learning and computer vision techniques, concentrating on categorizing and evaluating distinct batting shots from image or video material. A 2D CNN model obtained 91.5\% accuracy in classifying different cricket shots from images, indicating the promise of layered CNN architectures \cite{Fernandes2023Cricket}. Random Forest models using human body keypoints gathered using MediaPipe also exhibited good performance, with an F1-score of 87\%, and enabled similarity comparisons with professional players' shots for performance monitoring \cite{Devanandan2021Cricket}. Dey and Biswas \cite{Dey2024Shot-ViT} introduce Shot-ViT, a Vision Transformer-based model fine-tuned specifically for this task, which achieves a high validation accuracy of 92.58\% on CBSId, outperforming VGG19, ResNet50, I-AlexNet, and ViT-B32 by effectively capturing global context and long-range interdependence in images using self-attention processes. A 2025 study by Majumder et al. \cite{Majumder2025Cricket} finds 92.13\% accuracy for cricket shot recognition using image-based pose estimation. Although researchers conducted research using images to perform shot classification and performance analysis, but recent studies suggest insufficient role of image data for similar work. 
While focused on broad action recognition, Huang \cite{huang2022spatio} uses cricket examples to demonstrate that single-image approaches fail to convey underlying spatio-temporal information. Kanagal Sathyanarayana \cite{kanagal2025cricnet} notes restrictions such as occlusion, position fluctuation, and the absence of real-time dynamic analysis, and proposes video-based approaches to enhance biomechanical insight and robustness. 

Hoque et al. \cite{10441573} incorporate video-based cricket shot detection with a Long-term Recurrent Convolutional Network (LRCN) model, achieving 73\% accuracy. Semwal et al. Sen et al. \cite{sen2021cricshotclassify} achieve an accuracy of 93\% on the CricShot10 dataset. Similar efforts use 2D pose estimation on video data: a YOLOv8-based real-time pose-enhanced classification pipeline reaching 92.21\% \cite{Mobin2024CricPose360}, illustrating the benefit of pose-based data augmentation; and ResNet/2D-CNN approaches on video frames reaching around 90--91.5\% \cite{Fernandes2023Cricket}. 


Although 2D pose estimation approaches, such as those using OpenPose or MediaPipe, have achieved great accuracy in categorizing cricket shots by tracking joint positions in image planes, they lack depth information, limiting their capacity to completely capture the biomechanics of strikes \cite{siddiqui2023enhancing}. Recent studies applying deep learning with 3D pose estimation show promise in extracting more realistic motion patterns, enabling sharper classification of small shot variations and enhancing automated cricket shot recognition \cite{Datta2024Advancements}. Estimating 3D human poses from a monocular camera presents substantial challenges due to depth ambiguity and occlusion; nevertheless, recent advances in deep learning, particularly state-space and diffusion models, have markedly enhanced accuracy and robustness \cite{Guo2025A,J2025Deep}.

MeTRAbs (Metric-Scale Truncation-Robust Absolute 3D Human Pose Estimation) presents an innovative heatmap representation formulated exclusively in metric 3D space, facilitating the direct estimation of comprehensive, metric-scale 3D human poses without dependence on test-time distance information or anthropometric heuristics such as bone lengths. This method addresses the shortcomings of earlier 2.5D volumetric heatmaps that are aligned with image space and struggle with truncated images and scale ambiguity. MeTRAbs integrates 3D metric-scale heatmaps with 2D image-space heatmaps inside a differentiable architecture, attaining state-of-the-art performance on benchmarks such as Human3.6M, MPI-INF-3DHP, and MuPoTS-3D using a ResNet-50 backbone \cite{Sárándi2020MeTRAbs}. The technique has been effectively used in robotic applications employing fisheye cameras for proximate human--robot interaction and gesture recognition from a single camera \cite{Deshpande2024Novel}. Furthermore, MeTRAbs has refined pose-transfer methodologies by delivering more precise 3D posture inputs, augmenting the realism of produced images \cite{Watanabe2023Improvement}.

There is very little work on cricket shot classification and shot analysis using 3D pose estimation specifically. Some works  used MediaPipe pose detection to extract 3D landmarks, 
Randika et al. \cite{Randika2025Optimized} used four cameras with MediaPipe for real-time motion capture that classified batting shots with 95\% accuracy;
however, MediaPipe is not the most accurate model for 3D pose estimation, as discussed above. Deng et al. \cite{Deng2024Cricket} employ skeletal keypoint data extracted via OpenPifPaf as the basis for action classification, comparing a GCN approach (85\% accuracy) against a traditional CNN approach (60\% accuracy). 

Image-based classification is no longer state-of-the-art given how far computer vision has progressed; many video-based methods feed raw frames directly into models without explicit motion-feature extraction; and the small number of works that use 2D pose estimation are themselves limited by the absence of depth information. Existing evidence indicates that 3D pose estimation is likely to yield better shot detection and shot-analysis results, enabling useful biomechanical feedback for performance improvement. That's why, we propose a 3d pose-base classification mechanism to overcome most of these imitations and get a better result.

\begin{figure*}[t]
    \centering
    \includegraphics[width=0.95\textwidth]{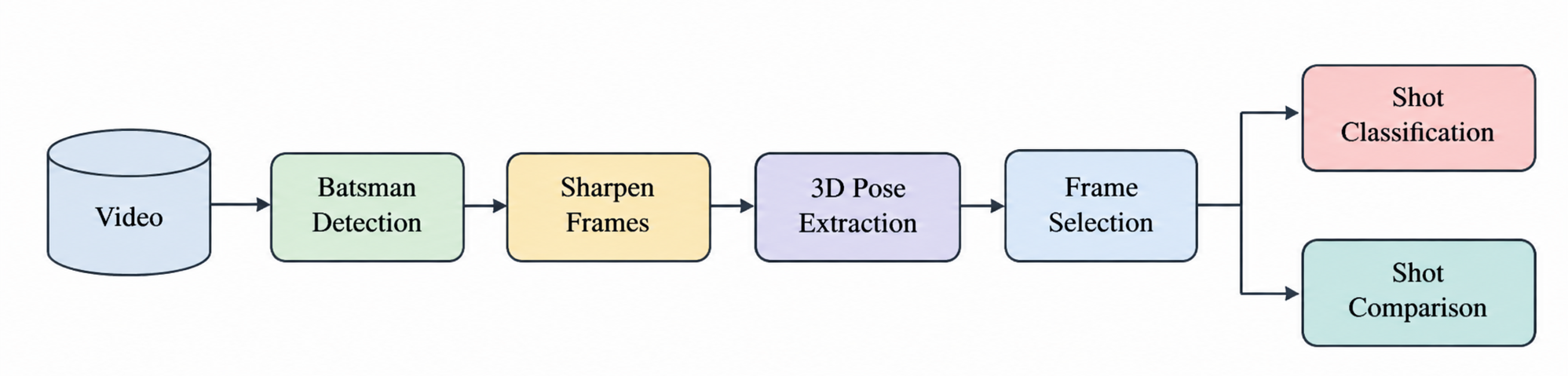}
    \caption{Pipeline of our proposed framework.}
    \label{fig:pipeline}
\end{figure*}

\section{Methodology}

We propose a video-based 3D pose analysis framework to classify cricket shots and compare performance between players. The methodology consists of five main stages, illustrated in Fig.~\ref{fig:pipeline}: video acquisition, batsman detection, 3D pose extraction, frame selection, and shot classification/comparison.

\subsection{Dataset Overview}
We used the KUCricShot Dataset \cite{10441573}, which consists of 1,278 cricket shot videos. The dataset contains raw short clips of the striker performing shots, mostly extracted from real-time match videos. The dataset contains four main shot types: drive, flick, defense, and pull, with no missing or erroneous labels. After careful manual inspection, 28 videos were excluded due to poor visual quality, including heavy occlusion, motion blur, or incomplete visibility of the striker. Thus, the final dataset contains 1,250 videos, comprising drive (335 videos, 26.8\%), flick (351 videos, 28.1\%), defense (318 videos, 25.4\%), and pull (246 videos, 19.7\%). 


\begin{figure}[t]
\centering
\begin{subfigure}{0.23\textwidth}
\includegraphics[width=\linewidth]{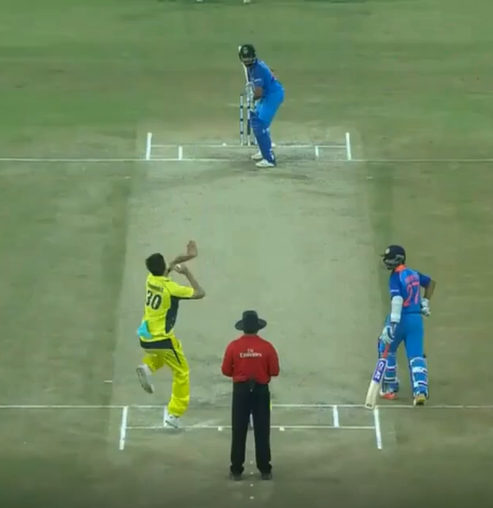}
\caption{Original frame}
\end{subfigure}
\hfill
\begin{subfigure}{0.23\textwidth}
\includegraphics[width=\linewidth]{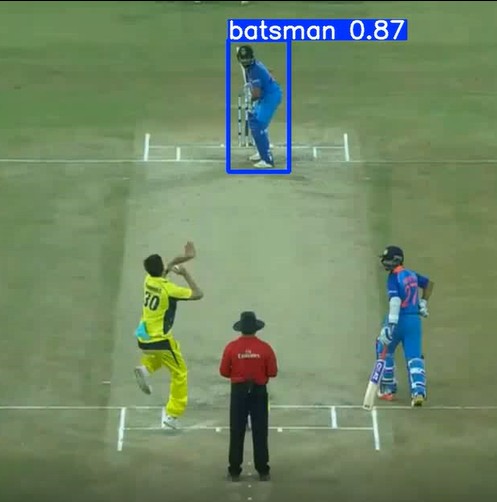}
\caption{YOLO detection}
\end{subfigure}
\caption{Comparison between original and YOLO-detected video frames.}
\label{fig:yolo}
\end{figure}

\subsection{Dataset Preprocessing}

To focus on the relevant region of each video frame, we used a YOLO model trained on the open-source Cricket Shots Dataset \cite{cricket-shots-wviff_dataset} to detect the batsman in every frame (Fig.~\ref{fig:yolo}). Simultaneously, all regions outside the YOLO bounding box are masked in black to suppress irrelevant information, ensuring that only the batsman remains visible for accurate 3D pose estimation.

After preprocessing, the frames are passed through MeTRAbs \cite{sarandi2021metrabs}. MeTRAbs generates a 3D skeletal representation for each frame in the form of SMPL30-style points, providing a consistent metric-scale joint representation across all frames and videos.

For each video, we uniformly sampled 15 frames. Experiments with 20 frames produced similar results; therefore, 15 frames were selected to reduce computational cost. Frame selection is based on the root mean squared error (RMSE) computed between consecutive frames, which reflects the degree of change in the batsman's movement over time.

To learn shot patterns that generalize across different players, body sizes, and camera angles, we applied a two-stage normalization process. First, all joint coordinates are re-centered around the pelvis (joint 0), so that each pose is represented relative to the player's center. Second, the centered pose is scaled using the average torso height across all frames, specifically, the pelvis-to-thorax distance between the pelvis (joint 0) and upper thorax (joint 12). This normalization reduces variations caused by differences in player height and distance from the camera.

After normalization, each shot is represented as a sequence of 15 frames, where each frame contains 30 body joint positions in 3D space using the SMPL30-style representation. This results in an input tensor of shape $(15 \times 30 \times 3)$. The four shot types are encoded as integer labels: drive ($y=0$), defense ($y=1$), flick ($y=2$), and pull ($y=3$).

\subsection{Shot Classification}
We used both machine learning and deep learning methods for shot classification. Five-fold stratified cross-validation was used for both approaches.

\subsubsection{Machine Learning Models}

We evaluated SVM, Random Forest, and XGBoost individually before combining them into a soft-voting ensemble. Each model's hyperparameters were tuned separately through cross-fold search to get the best performance out of it. The best hyperparameters obtained through cross-fold validation were $C=10$ and $\gamma=\text{scale}$ for SVM, $n\_estimators=200$, $max\_depth=\text{None}$, and $min\_samples\_split=2$ for Random Forest, and $n\_estimators=300$, $max\_depth=4$, $\eta=0.2$, $subsample=0.8$, and $colsample\_bytree=0.8$ for XGBoost.


\begin{figure*}[t]
\centering
\includegraphics[width=0.85\linewidth]{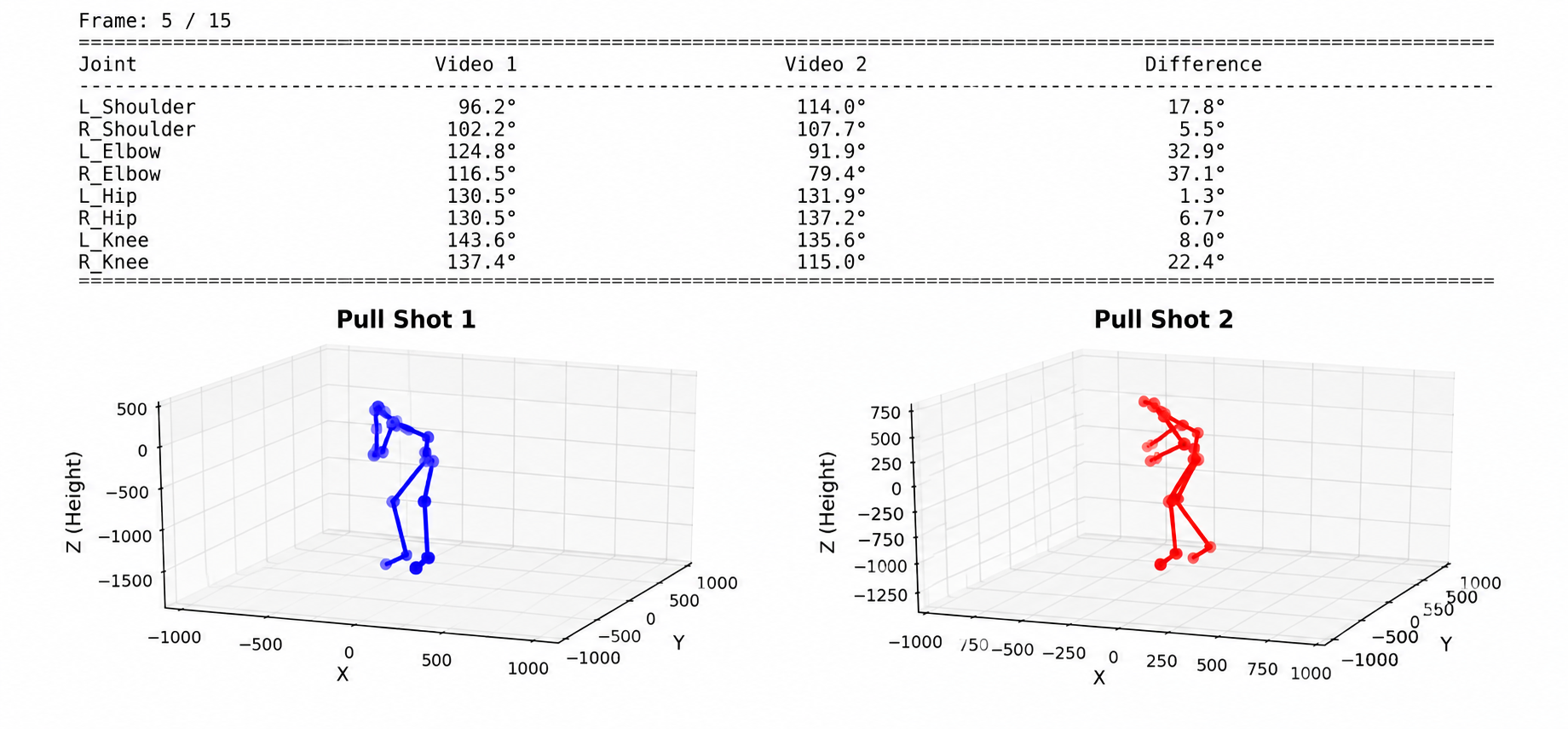}
\caption{Key joint comparison between two videos.}
\label{fig:jointcomp}
\end{figure*}
\subsubsection{Deep Learning Models}

For deep learning, we trained LRCN, TCN, and BiLSTM models to capture temporal patterns within the pose sequences. We also proposed an ensemble of LRCN and TCN to improve classification accuracy.

\paragraph{LRCN Model Architecture} Our LRCN model departs from the original formulation, which was built for image data — here it's adapted for pose-coordinate sequences instead. The $(15 \times 30 \times 3) $ pose tensor is reshaped to $(15 \times 90)$ and passed through a 128-neuron dense layer for feature extraction. A 1D convolutional layer (kernel size 3) then picks up local temporal patterns, while two bidirectional LSTM layers (64 and 32 units per direction) handle longer-range dependencies. A final 32-neuron dense layer feeds into a four-class softmax for classification.

\paragraph{TCN Model Architecture} The TCN uses dilated causal convolutions and residual connections to capture temporal patterns at different time scales. The reshaped pose sequence is first passed through a 128-neuron dense layer, followed by three residual TCN blocks with dilation rates of 1, 2, and 4. These progressively expand the receptive field, allowing the model to capture both short- and long-range motion patterns. Global average pooling is then used to summarize the sequence, followed by dense layers of 64 and 32 neurons and a final four-class softmax layer for classification.

Our proposed ensemble combines the TCN and LRCN by averaging their predicted probability distributions. The shot class with the highest averaged probability is selected as the final prediction. Combining the two models reduces prediction variance and improves robustness compared with either architecture alone.

Both LRCN and TCN were trained independently on each fold. Both models used the Adam optimizer with sparse categorical cross-entropy loss, a learning rate of 0.001, a batch size of 16, and a maximum of 200 epochs. Class weights were applied to reduce the effect of class imbalance. Dropout (30--40

We evaluated classification performance using accuracy, precision, recall, and F1-score, along with a $4 \times 4$ confusion matrix. The mean and standard deviation of accuracy across the five folds were reported to compare model performance.

\subsection{Shot Comparison and Parameter Analysis}

The framework also enables side-by-side comparison of key pose parameters across two shots. For each frame, joint angles, limb positions, and other relevant parameters were extracted. These parameters were compared across corresponding frames of the two shots to identify differences in movement execution. The observed differences are visualized in Fig.~\ref{fig:jointcomp}, which compares two pull-shot videos. The differences can also be represented through joint-angle variations to further assess an athlete's performance and provide feedback.


\section{Results and Discussion}
The experimental results show that both traditional machine learning and deep learning approaches were effective for cricket shot classification. The traditional machine learning ensemble achieved an average accuracy of \textbf{95.53\%}, while the deep learning models achieved higher performance overall. Among all evaluated models, the proposed ensemble of LRCN and TCN achieved the best result, with a mean accuracy\textbf{ of 97.68\% ± 0.99\%} across five-fold cross-validation.
\begin{table}[t]
\centering
\caption{Performance Comparison of Machine Learning Models}
\label{tab:ml_individual}
\small
\begin{tabular}{@{}lcccccc@{}}
\toprule
\textbf{Model} &
\textbf{Drive} &
\textbf{Pull} &
\textbf{Flick} &
\textbf{Defense} &
\textbf{Accuracy} \\
\midrule
SVM &
0.96 & 0.95 & 0.95 & 0.91 & 0.94\\

Random Forest &
0.94 & 0.97 & 0.91 & 0.87 & 0.92\\

XGBoost &
0.98 & 0.97 & 0.93 & 0.91 & 0.94\\

Ensemble &
0.96 & 0.97 & 0.95 & 0.93 & 0.95\\

\bottomrule
\end{tabular}
\end{table}
\subsection{Machine Learning Model Performance}
Table~\ref{tab:ml_individual} summarizes the performance of individual traditional machine learning models on the test dataset. SVM and XGBoost led the individual classifiers with 94\% accuracy each; Random Forest followed at 92\%. XGBoost's macro-averaged F1-score was higher still, pointing to better balance across shot classes. The ensemble did better than any single model, averaging 95.53\% accuracy with just 1.26\% standard deviation — consistent performance across data splits.

\subsection{Deep Learning Model Performance}
The deep learning models also performed well in capturing the temporal patterns of cricket shots. Table~\ref{tab:dl_individual} presents a comparison of the three models. LRCN demonstrated consistent performance across folds, with a mean accuracy of 96.43\% and a standard deviation of 1.81\%. TCN achieved a mean accuracy of 95.79\%, with lower variability (1.27\% standard deviation) compared with LRCN. Although BiLSTM achieved a competitive overall accuracy of 92.46\%, it showed greater class-specific variation, with strong performance on flick and pull shots, both achieving F1-scores above 94\%, but weaker performance on defense.

Table~\ref{tab:arch_comparison} provides a comparison of the three architectures across the key evaluation metrics. 

LRCN came out on top for overall accuracy and macro F1-score. TCN, meanwhile, was the most stable across folds (1.27\% standard deviation) and handled drive shots best.


\begin{table}[t]
\centering
\caption{Performance Comparison of Deep Learning Models}
\label{tab:dl_individual}
\small
\begin{tabular}{@{}lccccc@{}}
\toprule
\textbf{Model} &
\textbf{Drive} &
\textbf{Pull} &
\textbf{Flick} &
\textbf{Defense} &
\textbf{Accuracy} \\
\midrule

LRCN &
0.9632 & 0.9798 & 0.9712 & 0.9464 & 0.9643 \\

TCN &
0.9689 & 0.9741 & 0.9585 & 0.9333 & 0.9579 \\

BiLSTM &
0.9385 & 0.9592 & 0.9429 & 0.8676 & 0.9246 \\

Ensemble &
0.9792 & 0.9623 & 0.9771 & 0.9919 & 0.9768 \\
\bottomrule
\end{tabular}
\end{table}


\begin{table}[t]
\centering
\caption{Shot Misclassification Summary}
\label{tab:error_analysis}
\small
\begin{tabular}{@{}lccc@{}}
\toprule
\textbf{True} & \textbf{Predicted} & \textbf{Count} & \textbf{Error \%} \\
\midrule
Defense & Drive & 12 & 41.4\% \\
Defense & Flick & 6 & 20.7\% \\
Flick & Defense & 3 & 10.3\% \\
Pull & Flick & 2 & 6.9\% \\
Drive & Flick & 2 & 6.9\% \\
Flick & Pull & 1 & 3.4\% \\
Flick & Drive & 1 & 3.4\% \\
Defense & Pull & 1 & 3.4\% \\
Drive & Defense & 1 & 3.4\% \\
\midrule
\textbf{Total} & & \textbf{29} & \textbf{100.0\%} \\
\bottomrule
\end{tabular}
\end{table}


\begin{table}[!t]
\centering
\caption{Deep Learning Model Performance Comparison}
\label{tab:arch_comparison}
\small
\begin{tabular}{@{}lcccc@{}}
\toprule
\textbf{Metric} & \textbf{LRCN} & \textbf{TCN} & \textbf{BiLSTM} & \textbf{Best} \\
\midrule
Overall Accuracy & 96.43\% & 95.79\% & 92.46\% & LRCN \\
Drive F1-Score & 96.32\% & 96.89\% & 93.85\% & TCN \\
Defense F1-Score & 94.64\% & 93.33\% & 86.76\% & LRCN \\
Flick F1-Score & 97.12\% & 95.85\% & 94.29\% & LRCN \\
Pull F1-Score & 97.98\% & 97.41\% & 95.92\% & LRCN \\
\bottomrule
\end{tabular}
\end{table}




Table~\ref{tab:prior_comparison} compares different model configurations for cricket shot classification on the similar type dataset. The results show that combining pose information with an ensemble approach significantly improves accuracy over raw-video-based methods, yielding a \textbf{+4.7}-point improvement over the 2D-pose approach and a \textbf{+24.4}-point improvement over the raw-video approach.
The ensemble achieved a mean accuracy of \textbf{97.68\% $\pm$ 0.99\%}. 
a substantial improvement over the individual architectures, and outperformed the better individual model.


\begin{table}[t]
\centering
\caption{Comparison of our model with previous works}
\label{tab:prior_comparison}
\small
\begin{tabular}{@{}p{0.24\linewidth}p{0.50\linewidth}p{0.14\linewidth}}
\toprule
\textbf{Model Type} & \textbf{Key Characteristics} & \textbf{Accuracy} \\
\midrule
LRCN with raw video\cite{10441573} & Direct pixel processing, appearance-based & 73.0\% \\
LRCN with 2D pose\cite{Mobin2024CricPose360} & Skeletal representation, appearance-invariant & 93.0\%  \\
\textbf{Our proposed model} & \textbf{3D pose,(TCN + LRCN) ensemble} & \textbf{97.68\%} \\
\bottomrule
\end{tabular}
\end{table}

\subsection{Ensemble Error Analysis}
Detailed analysis of the 29 misclassified samples (2.32\% error rate) across all folds is summarized in Table~\ref{tab:error_analysis}. The error analysis shows that most mistakes occurred between shots with similar body movements. Drive and defense produced the highest confusion, with 13 cases representing 44.8\% of all errors; this confusion may occur because both shots use a similar forward posture and bat path. Defense and flick were confused in nine cases, accounting for 31.0\% of the errors, as these shots can appear similar when the batter uses a small, controlled bat movement. Only three errors occurred between flick and pull, showing that these shots were generally well separated, and no confusion was found between drive and pull because their body positions and movement patterns are clearly different. Overall, the model performed well, but had difficulty distinguishing shots with subtle differences in posture and follow-through.

\subsection{Discussion}

The experimental results demonstrate that our proposed pose-based approach can effectively classify cricket shots from 3D skeletal sequences. Among the evaluated models, the ensemble of LRCN and TCN achieved the highest performance. This improvement suggests that the two architectures capture complementary temporal information from the same pose sequence. LRCN is effective at learning sequential dependencies, while TCN captures temporal patterns at different scales. Combining these representations also improved the classification of the more challenging defense shots, increasing the F1-score from 0.9464 for LRCN to 0.9623 for the ensemble.

The use of 3D skeletal information provides an additional advantage by focusing on the batsman's body movement rather than appearance-based information. This makes the classification less dependent on background and lighting conditions and also provides a representation that can be interpreted in terms of body movement. Such a representation is particularly useful for a coaching system, where understanding differences in movement between shot attempts is as important as identifying the shot type.

Despite the promising results, the study has several limitations. The dataset contains 1,250 samples and covers only four shot categories, which limits the range of batting actions represented in the experiments. In addition, the current system relies on pre-segmented shots, so automatic identification of shot boundaries is not yet addressed. Although GAN-based synthetic data generation could increase the number of training samples, it was not used because generated poses may not always represent realistic batting movements and could introduce patterns that are not present in real data. Therefore, the study prioritizes the use of real samples for maintaining the reliability of the evaluation.

\section{Conclusion and Future Work}

This work presents a 3D pose-based framework for cricket shot classification, combining MeTRAbs pose estimation with a LRCN--TCN ensemble. Our proposed ensemble reached 97.68\% mean accuracy, showing that combining these two temporal models captures complementary information from 3D skeletal sequences. Beyond classification, the framework can compare different attempts at the same shot by tracking changes in joint movement, which lays groundwork for more detailed batting analysis.

Several directions remain open. The dataset needs more real-world samples and a wider range of shot categories, and automatic shot segmentation would remove the current dependence on pre-segmented clips. A natural next step is a quality assessment module that evaluates specific aspects of batting technique rather than just classifying the shot. Pairing this with a large language model (LLM) could turn raw pose analysis into plain-language feedback, useful for summarizing performance trends or helping coaches make training decisions.

The same joint-tracking approach also opens a physiotherapy-adjacent use case: flagging unusual or repetitive movement patterns that might warrant a closer look from a professional. This would function purely as a decision-support aid, not a diagnostic tool. Taken together, pose estimation, temporal modeling, and language-based feedback point toward a more complete AI-assisted cricket coaching system.

\bibliographystyle{IEEEtran}
\bibliography{references}

\end{document}